\documentclass{elsarticle}

\usepackage{lineno,hyperref}
\modulolinenumbers[5]

\journal{Future Generation Computer Systems}

\newtheorem{exmp}{Example}[section]

\usepackage{caption}	
\usepackage{graphicx}
\usepackage{amsmath}
\usepackage{subfig}

\begin{document}
	
\begin{frontmatter}
	
	\title{Context-driven Active and Incremental Activity Recognition}
	
	\author{Gabriele Civitarese\corref{mycorrespondingauthor}}
	\ead{gabriele.civitarese@unimi.it}
	\author{Riccardo Presotto\corref{mycorrespondingauthor2}}	
	\ead{riccardo.presotto@studenti.unimi.it}
	\author{Claudio Bettini\corref{mycorrespondingauthor2}}
	\address{EveryWare Lab, Dept. of Computer Science\\ Universit\`a degli Studi di Milano, Via Celoria 18, Milan, Italy}
	\ead{claudio.bettini@unimi.it}


	\cortext[mycorrespondingauthor]{Corresponding author}

	
	\begin{abstract}
Human activity recognition based on mobile device sensor data has been an active research area in mobile and pervasive computing for several years. While the majority of the proposed techniques are based on supervised learning, semi-supervised approaches are being considered
to significantly reduce the size of the training set required to initialize the recognition model. These approaches usually apply self-training or active learning to incrementally refine the model, but their effectiveness seems to be limited to a restricted set of  physical activities.  
We claim that the context which surrounds the user (e.g., semantic location, proximity to transportation routes, time of the day) combined with common knowledge about the relationship between this context and human activities could be effective in significantly increasing the set of recognized activities including those that are difficult to discriminate only considering inertial sensors, and the ones that are highly context-dependent.

In this paper, we propose CAVIAR, a novel hybrid semi-supervised and knowledge-based system for real-time activity recognition. 
Our method applies semantic reasoning to context data to refine the prediction of a semi-supervised classifier. The context-refined predictions are used as new labeled samples to update the classifier combining self-training and active learning techniques. 
Results on a real dataset obtained from $26$ subjects show the effectiveness of the context-aware approach both on the recognition rates and on the number of queries to the subjects generated by the active learning module.
In order to evaluate the impact of context reasoning, we also compare CAVIAR with a purely statistical version, considering features computed on context data as part of the machine learning process.

	\end{abstract}
	
	
\end{frontmatter}

\linenumbers

\section{Introduction}
\label{sec:intro}
The evolution of mobile computing in the last decades allowed to develop intelligent applications that continuously monitor our daily activities to provide context-aware services~\cite{lara2013survey}. The majority of activity recognition algorithms in the literature rely on supervised machine learning to infer the most likely performed activities by analyzing inertial sensors data~\cite{kwapisz2011activity}. One of the major drawbacks of those solutions is the cost of collecting the amount of labeled data required to reach a high recognition rate. Moreover, standard classifiers are trained once with available data, and the recognition model cannot evolve over time.
To overcome these issues, semi-supervised and incremental approaches for activity recognition have been proposed~\cite{abdallah2018activity}.
Those methods only require a small amount of training data to initialize the recognition model, while techniques like co-learning, self-learning or active learning are used to assign labels to unlabeled sensor data~\cite{hossain2017active,abdallah2015adaptive,longstaff2010improving,stikic2008exploring}. 

While the majority of semi-supervised methods showed to be effective on classifying few physical activities (e.g., walking, running, biking, etc.), their effectiveness on more complex and context-dependent activities is still unclear. Moreover, discriminating those activities which have similar motion patterns is still problematic. For instance, activities like \emph{walking} and \emph{taking the stairs}, or \emph{standing} and \emph{taking the elevator} are easily confused between them by purely statistical methods based on inertial sensors. 

Considering the context which surrounds the user could be valuable information to mitigate these issues~\cite{liao2006location,RiboniB11}. Indeed, considering a rich description of the user's context (e.g., semantic location, weather, traffic condition, speed, etc.) has the potential to enable the recognition of a wide set of activities which are a) highly dependent to the current context and b) difficult to recognize only considering inertial sensors. However, semi-supervised approaches rely on a small set of labeled data, while activities can be performed in a large number of possible context conditions. 
For this reason, directly using context-data as additional features in the machine learning process may not be as effective as expected. 

In this paper, we consider this problem and we propose CAVIAR a Context-aware ActiVe and Incremental Activity Receognition system which combines semi-supervised learning and semantic context-aware reasoning. 
A real-time incremental classifier is in charge of analyzing inertial sensors data obtained from mobile devices to provide probability distributions over the possible activities.
A knowledge-based reasoning engine is then used to exclude from the statistical predictions those activities which are highly unlikely considering context-data. The system's output is the most likely activity from the resulting context refinement.

Following the semi-supervised approach, context-refined predictions are used to update the incremental classifier. When CAVIAR is confident about the refined prediction, it is provided as a new labeled sample to the incremental classifier. On the other hand, when it is not sufficiently confident, CAVIAR asks the ground truth to the user and uses the answer also as a new labeled sample to refine the machine learning model. 

In order to evaluate CAVIAR, we acquired a large dataset of inertial sensor data and rich contextual information. 
Results on this dataset show that context-refinement is effective in improving the recognition rate and, at the same time, triggering a significantly lower number of queries.

The contributions of our work are the following: 
\begin{itemize}
    \item We propose a novel method to combine context-aware reasoning with semi-supervised learning method for activity recognition. 
    \item We acquired a realistic labeled dataset of activities performed by $26$ subjects, collecting data both from inertial sensors data and several sources of context.
    \item We performed an extensive evaluation of our approach on this dataset showing the crucial role of context-data and structured knowledge in improving semi-supervised activity recognition.
    \item We show that using knowledge-based reasoning on context-data not only allows reaching higher recognition rates, but also to obtain a significantly lower number of queries in active learning compared to using context-data as additional features in the machine learning process. 
\end{itemize}

The rest of the paper is organised as follows. Section~\ref{sec:related} discusses related work. Section~\ref{sec:system} describes the overall architecture of CAVIAR. Section~\ref{sec:incremental} presents the CAVIAR method in details. Section~\ref{sec:evaluation} reports the experimental results. Section~\ref{sec:discussion} discusses strengths and limitations of CAVIAR. Finally, Section~\ref{sec:conclusion} concludes the paper.

\section{Related work}
\label{sec:related}
The recognition of physical activities using commonly available mobile devices (e.g., smartphones and smartwatches) is a widely explored research area~\cite{lara2013survey,shoaib2015survey}. The majority of approaches in the literature rely on supervised methods to infer activities from inertial sensors data~\cite{kwapisz2011activity, gyorbiro2009activity, sun2010activity, bao2004activity, BullingBS14}. While these methods allow reaching high recognition rates, the acquisition of a wide labeled dataset of activities is costly and often unfeasible~\cite{CookFK13}.

In order to overcome these issues, few works proposed unsupervised learning techniques~\cite{kwon2014unsupervised, trabelsi2013unsupervised, lee2009unsupervised}. 
These methods aim to find activity patterns from unlabeled data with data mining techniques.
However, the discovery of those patterns requires the acquisition of a large dataset of unlabeled data. 
Moreover, a certain amount of labeled data are still required in order to reliably associate each cluster with its corresponding activity class.
Knowledge-based methods coupled with unsupervised learning have been proposed to automatically label activity traces~\cite{chen2014ontology}. While this methodology is suitable for smart-home activity recognition using environmental sensors, it is not directly applicable to the recognition of activities using the sensor data of mobile devices.  

In order to combine the strengths of both supervised and unsupervised approaches, semi-supervised learning methods for activity recognition have been proposed~\cite{abdallah2018activity, stikic2008exploring, guan2007activity, longstaff2010improving}. Those techniques use small labeled training sets to initialize the model, which is continuously enhanced using unlabeled data. In the literature, the semi-supervised approaches which have been mainly considered for activity recognition are self-learning, co-learning, and active learning.
Self-learning methods exploit the starting small training set to classify unlabeled data~\cite{longstaff2010improving}. The most confident predictions are hence used to update the classifier. Co-learning involves multiple classifiers trained on different data perspectives. Those classifiers collaboratively improve their models exploiting their prediction's confidences~\cite{lee2014activity, guan2007activity}.
Differently from self-learning and co-learning, active learning requires explicit feedback from the users in order to obtain labels for the most informative data (i.e., data where the classifier is uncertain about the performed activity)~\cite{hoque2012aalo, miu2015bootstrapping, abdallah2015adaptive, hossain2017active}.
Active learning proved to be particularly effective for semi-supervised activity recognition. However, for the sake of usability the number of triggered queries should be low.

Existing semi-supervised activity recognition methods in mobile computing are mainly based on the analysis of inertial sensors data to recognize a restricted number of physical activities~\cite{miu2015bootstrapping, abdallah2015adaptive, lee2014activity, huynh2006towards}. Differently from those methods, we consider the context which surrounds the user to continuously update an incremental classifier through a combination of self-learning and active learning. This allows us to significantly extend the set of recognized activities and, at the same time, to better discriminate those activities which have similar motion patterns.

Even if context reasoning for activity recognition has been mainly investigated for smart-home environments~\cite{rodriguez2014survey} and computer vision based systems~\cite{akdemir2008ontology}, its application to mobile computing applications is not completely new~\cite{yurur2016context}.
The combination of machine learning and context-aware ontological reasoning for activity recognition with mobile devices was firstly explored in~\cite{RiboniB11}. In that work, the output of a statistical classifier is refined considering the user's semantic location. In~\cite{saguna2013complex} rich contextual information is used to improve activity recognition with a multi-layer approach.  Differently from those methods, our system is semi-supervised and hence it only requires a small labeled dataset in order to be initialized. Moreover, our approach takes advantage of context-aware reasoning to continuously update and personalize the activity model.


\section{{CAVIAR} system overview}
\label{sec:system}
\begin{figure}[h!]
	\centering
	\includegraphics[width=0.7\textwidth]{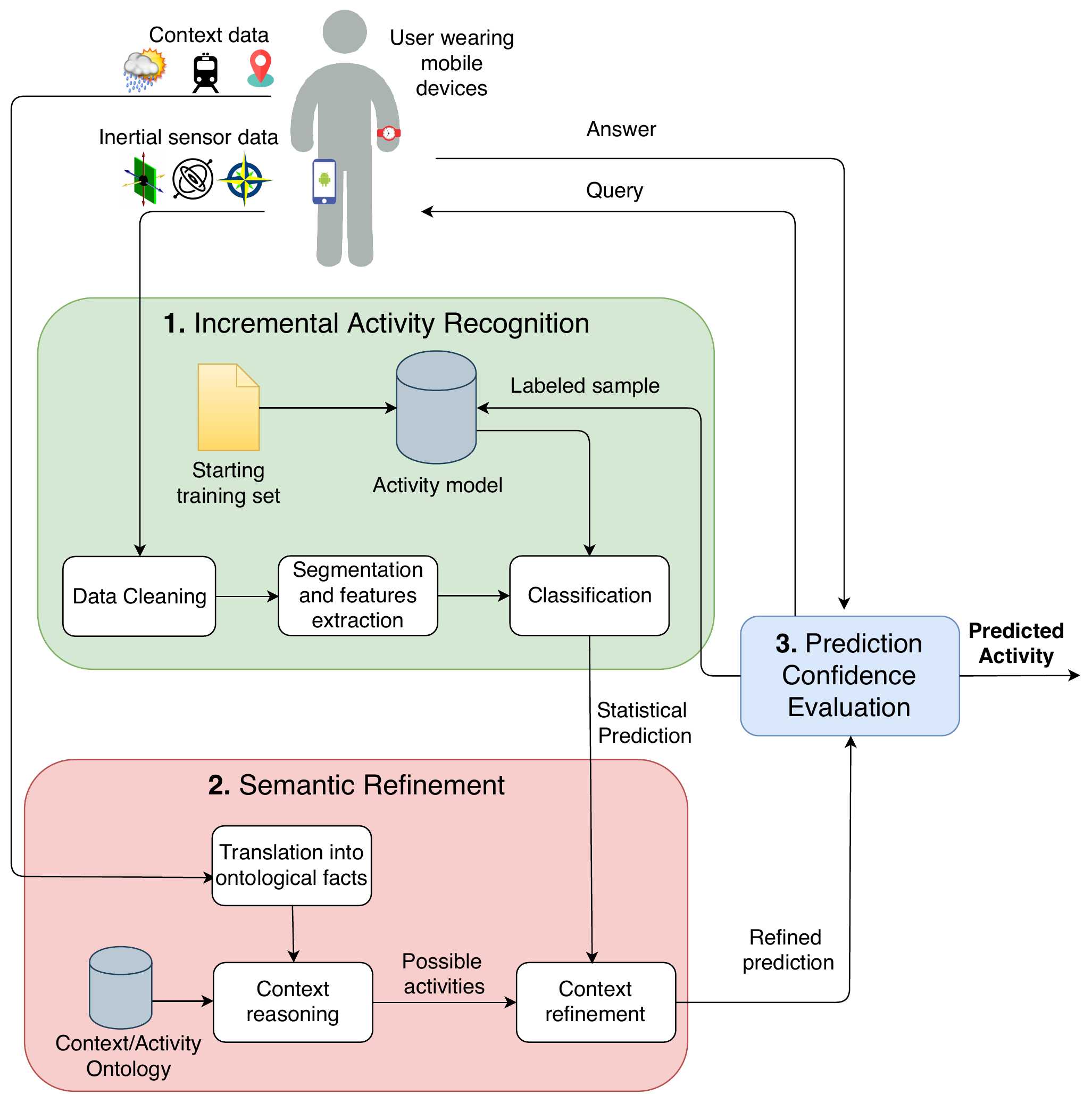}
	\caption{Overall architecture of our system}
	\label{fig:arch}
\end{figure}

The general architecture of our system is shown in Figure~\ref{fig:arch}.
The user's mobile devices continuously acquire data from different sources. One one hand, inertial sensors (e.g., accelerometer, magnetometer and gyroscope) are in charge of continuously streaming data about the physical movements of the user. On the other hand, data from other built-in sensors and devices (e.g., GPS) in combination with publicly available web services (e.g., weather service) are used to obtain data about user's context. 
It is important to note that ``\emph{context}'' is a very broad term which in the literature is used to model users situations at several level of abstractions~\cite{PMCsurvey09}. For instance, even the user's performed activity can be considered as high-level contextual information. 
In this paper, with context data we mainly indicate the information about the environment which surrounds the user.
Examples of such context data are user's current semantic location, his/her proximity to transportation routes, the current weather, the time of the day, the day of the week, etc.

Our hybrid semi-supervised and context-aware activity recognition algorithm is mainly divided into three steps.
First, the stream of raw inertial sensors readings is processed by the \textsc{Incremental Activity Recognition} module. This module first applies pre-processing methods like data cleaning, segmentation, and feature extraction to raw inertial sensors data. Then, a semi-supervised classifier associates to each feature vector a probability distribution over the possible activities. It is important to note that the activity model needs to be initialized with a small labeled training set in an offline phase. 

The \textsc{Incremental Activity Recognition} module does not take into account context data. The main motivation is that semi-supervised methods rely on a rather small set of labeled data, while activities can be performed in a wide variety of different contexts. While it is feasible for a classifier to discriminate different motion patterns even with few labeled samples, learning their correlations with all the possible context conditions may be problematic. This is especially true when considering a wide set of activities. On the other hand, common-sense knowledge can be used to model the relationships between activities and contexts 
(e.g., it is unlikely that a user is using an elevator while she is in the city park).
Hence, in the second step of our algorithm, the \textsc{Semantic Refinement} module applies knowledge-based reasoning to context data in order to exclude from the semi-supervised prediction those activities which are not consistent with the current context. In particular, context data needs to be pre-processed and translated into high-level facts, which are mapped to an ontology that models activities and contexts. 
Knowledge-based reasoning is then applied to evaluate which activities are \emph{context-consistent}.
The output of the \textsc{Semantic Refinement} module is a probability distribution over the \emph{context-consistent} activities (which we will refer as \emph{refined prediction}). 

The third and the last step of our method consists of using the refined prediction to update the semi-supervised activity model. In particular, the \textsc{Prediction Confidence Evaluation} module evaluates the system's confidence on the refined prediction. If the confidence is sufficiently high, the refined prediction is provided as a new labeled example to our incremental classifier. Otherwise, a query is triggered to the user in order to obtain the ground truth about the current activity, in order to update the recognition model accordingly. 

In our architecture, the semi-supervised activity model is stored on a server and shared between all the participating users, which can collaboratively update it using our context-driven semi-supervised framework. 

\section{Methodology}
\label{sec:incremental}

In this section, we describe in details the different modules of our system introduced in Section~\ref{sec:system}.

\subsection{Incremental activity recognition}
\label{subsec:incrementalactivityrecognition}

The \textsc{Incremental Activity Recognition} module relies on a semisupervised classifier to derive from inertial sensors data a set of candidate activities performed by the user in real-time. In particular, the sensors' signal is pre-processed and segmented in order to extract feature vectors.
For each feature vector, the \textsc{Incremental Activity Recognition} module outputs a probability distribution over the considered activities.
As we will explain later in this section, this probability distribution is refined by the \textsc{Semantic Refinement} module which is then used by the \textsc{Prediction Confidence Evaluation} module to update the incremental classifier with new labeled samples.

\subsubsection{Segmentation, feature extraction and classification}
\label{subsubsec:fv}

In the following, we describe the data flow of inertial sensors data for activity recognition.
Since a user may carry multiple mobile devices (e.g., a smartphone and a smartwatch), it is first necessary to temporally align their raw sensor data streams. In our experimental setup we considered for each device the data streams from accelerometer, magnetometer and gyroscope.
We apply a median filter to reduce the noise in each signal from the streams.
Then, we segment the streams of aligned sensor data. Each segment is defined as the set of inertial sensor data acquired during a specific time window of $n$ seconds.
Each segment starts the next second with respect to the end of the previous segment, hence segments are contiguous and non-overlapping. 
The length $n$ is the same for all segments, and it must be chosen carefully according to the complexity of the considered activities~\cite{banos2014window}. 

From each segment, we extract a wide set of statistical features which are well-known in the activity recognition literature~\cite{lara2013survey}. 
In particular, for each axis of each inertial sensor, we extract: \emph{average}, \emph{variance}, \emph{standard deviation}, \emph{median}, \emph{mean squared error}, \emph{kurtosis}, \emph{symmetry}, \emph{zero-crossing rate}, \emph{number of peaks}, \emph{energy} and \emph{difference between maximum and minimum}.
Finally, for each inertial sensor we compute the \emph{pearson correlation} for each combination of its axes and \emph{magnitude} on all of its axes.
Hence, given $k$ 3-axis inertial sensors equipped in the user's mobile devices, we compute $k \times 37$ features.
We also apply standardization to each feature in order to further improve the recognition rate~\cite{guyon2006introduction}.
\begin{exmp}
Consider a user which carries a smartphone and a smartwatch, both equipped with 3-axis accelerometer, gyroscope and magnetometer.
Hence, the overall number of inertial sensors is $6$. In this scenario, our feature extraction mechanism would compute, for each segment, $6 \times 37 = 222$ features.
\end{exmp}

For each feature vector $fv$ computed from a segment $s$, the incremental classifier $h$ outputs a probability distribution over the set of considered activities $\mathbf{A} = \{A_1, A_2, \dots, A_m\}$:
$$ h(fv) = \langle p_1, p_2, \dots, p_m \rangle $$
where $0 \leq p_i \leq 1$ is the probability $P(A_i | s)$ that the segment $s$ was generated by the activity $A_i$, $\sum_{i}^{m} p_i = 1$, and $m = |\mathbf{A}|$.
The probability distribution $h(fv)$ is forwarded to our \textsc{Semantic Refinement} module which will refine it using contextual information.

\subsubsection{Activity model bootstrap}
\label{subsec:bootstrap}

A crucial aspect of our semi-supervised framework is the activity model initialization. 
Indeed, without a proper bootstrap mechanism the semi-supervised model would have to discover each activity ``on-the fly'', with a negative impact on the recognition rate. Hence, we initialize the semi-supervised model acquiring $t$ seconds of labeled data for each activity in order to obtain a balanced labeled dataset.
Clearly, the parameter $t$ has a high impact both on the recognition accuracy and on the number of queries triggered to the users.
However, as we motivated in the introduction, it is unfeasible to obtain a wide labeled dataset (i.e., choosing a high value of $t$).
The choice of $t$ mainly depends on the number of considered activities and their complexity.
In order to reduce the effort of acquiring and annotating data, it is hence important to use a small value of $t$ which allows achieving a reasonable recognition accuracy. We adopt an empirical approach to determine the value of $t$, as illustrated in Section \ref{sec:evaluation}.

\subsection{Semantic Refinement}
\label{subsec:ontology}

The \textsc{Semantic Refinement} module is in charge of analyzing the context which surrounds the user to refine the prediction $h(fv)$ obtained by the \textsc{Incremental Activity Recognition} module. In order to achieve this task, this module relies on an ontology which models the relationships between contexts and activities. In particular, ontological reasoning is applied to exclude from the statistical prediction those activities which are unlikely considering the current context.
In the following, we describe in details our context-aware semantic reasoning mechanism. 

\subsubsection{Activity and Context ontology}
\label{subsec:ontologydescription} 
\begin{figure}[h!]
	\centering
	\subfloat[An excerpt of context hierarchy]{\includegraphics[width=0.35\columnwidth]{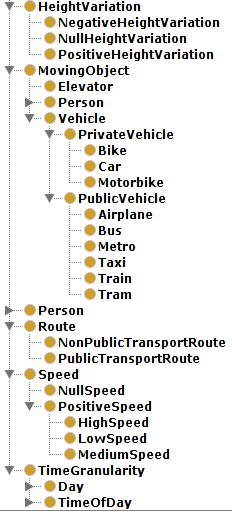}
		\label{fig:ontologycontext}}
	\hfil
	\subfloat[An excerpt of symbolic locations hierarchy]{\includegraphics[width=0.40\columnwidth]{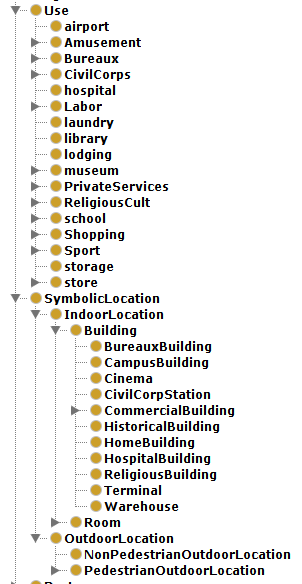}
		\label{fig:ontologylocation}}
	\caption{Excerpts of our ontology}
	\label{fig:ontologyexcerpt}
\end{figure}

In order to enable semantic reasoning with context data for activity recognition, we extended the \emph{ActivO} ontology~\cite{RiboniB11}.
That ontology defines a wide set of activities, semantic locations, artifacts (e.g., used by the user or part of the semantic locations), user's postures, time granularities (e.g., day of the week, time of the day) and environmental information (e.g., temperature and light conditions). Details about \emph{ActiveO}'s implementation can be found in~\cite{RiboniB11}. We took advantage of the Prot\'eg\'e tool \footnote{\url{https://protege.stanford.edu/}} to extend \emph{ActivO} with several new activities, contextual data and their relationships. An example of those entities are shown in Figure~\ref{fig:ontologyexcerpt}.
Our ontology considers several sources of context data: \emph{user's semantic place}, \emph{user's recent route}, \emph{weather conditions}, \emph{proximity to public transportation stops and routes}, \emph{surrounding traffic condition}, \emph{user's height variations}, \emph{user's speed}, \emph{surrounding light}, \emph{environment's noise level} and \emph{temporal context (e.g., time of the day, day of the week, month, \dots)}. 
Figure~\ref{fig:ontologycontext} shows a portion of those context data modeled in our ontology, while Figure~\ref{fig:ontologylocation} focuses on the set of considered semantic locations, including the ones classified by Google Places API~\footnote{\url{https://developers.google.com/places/supported_types}}. It is important to note that we distinguish symbolic locations/buildings (and their characteristics) from their use. This allows us to better model activities related to symbolic locations.

\begin{figure}[h!]
	\centering
	\subfloat[Definition of the activity ``taking stairs'']{\includegraphics[width=0.49\columnwidth]{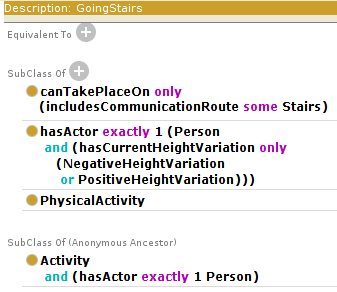}
		\label{fig:goingstairs}}
	\hfil
	\subfloat[Definition of the activity ``moving by car'']{\includegraphics[width=0.49\columnwidth]{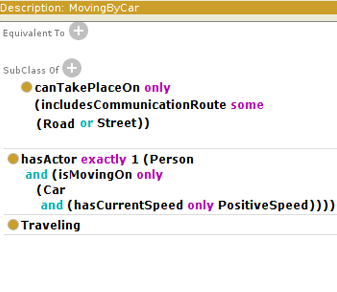}
		\label{fig:movingbycar}}
	\caption{Examples of activity definitions in our ontology}
	\label{fig:ontexamples}
\end{figure}

Due to the intrinsic open-world assumption of ontological reasoning, we explicitly state the necessary conditions which make activities possible or not possible in a given context. As we will explain later, such constraints are necessary to enable our context-aware refinement which is based on \emph{consistency} reasoning.
For instance, the activity \texttt{TakingStairs} (Figure~\ref{fig:goingstairs}) should take place at a location which may have stairs and the person should have a non-negative height variation. Another example is the activity \texttt{MovingByCar} (Figure~\ref{fig:movingbycar}): our ontology enforces that it should take place in an outdoor location which includes a road or a street, and that the car's speed should be positive.

\subsubsection{Context reasoning and refinement}
\label{subsec:consistency}

For each activity candidate $A$ in the prediction $h(fv)$, we use ontological reasoning to determine whether $A$ is consistent or not with the surrounding context.  
First, CAVIAR adds an axiom to represent an instance of \texttt{Person} which identifies the subject wearing the mobile devices.
Then, it is necessary to instantiate relationships between the individual and context data.
For this reason, context data collected by the mobile devices need to be mapped to ontological concepts.
The majority of context-data has a mapping one-to-one with ontological entities.
However, scalar values need to be discretized and mapped to the classes covered by the ontology. 
Finally, since we want to test the consistency of an activity $A$ with respect to the current context, CAVIAR will add an axiom which states that the user is performing $A$. We define an activity $A$ \emph{context-consistent} when the axioms created with the observed data as described above are consistent with respect to the domain knowledge. Note that the consistency check involves reasoning that is automatically performed in the logic used to specify the ontology (the specific language and reasoner used in CAVIAR are reported in Section \ref{sec:evaluation}).

\begin{exmp}
	\label{exmp:running}
	Bob is using CAVIAR. When the context reasoning task is triggered, \texttt{Person(Bob)} is added as a fact.
	Then, the context data gathered by the mobile devices is analyzed in order to expand the set of facts.
	Suppose that some Web service provided the information that Bob is in a park and that the speed value obtained by the GPS sensor is $10$ km/h. First, we have to instantiate the individuals for the context data: \texttt{Park(place)}, \texttt{MediumSpeed(speed)}. Note that the raw speed value obtained by the GPS has been discretized in order to be mapped to an ontological concept.
	Then, the relationships between Bob and context data are added as facts:
	 \texttt{hasCurrentSymbolicLocation(Bob, place)}, \texttt{hasCurrentSpeed(Bob, speed)}. Finally, in order to test whether the activity \texttt{Running} is context-consistent, we add the axioms \texttt{Running(currentActivity)} and \texttt{isPerforming(Bob, currentActivity)}. The consistency of the set of facts with respect to the domain knowledge will determine if the running activity is consistent according to the current Bob's context.
\end{exmp}

Given the current context $C$ and the marginal probabilities obtained by the semi-supervised classifier $h(fv) = \langle p_1, p_2, \dots, p_m \rangle$, the goal of context refinement is to exclude those activities which are not \emph{context-consistent} according to $C$.
For each activity class $ac_i$ such that $p_i>0$, we compute its consistency according to context $C$ as explained above. Each activity which is not \emph{context-consistent} is removed from the probability vector. The refined vector is finally normalized in order to preserve the properties of a probability distribution.
The output is a new refined probability vector $\langle r_1, r_2, \dots, r_c \rangle$ such that each $A_i$ is a \emph{context-consistent} activity according to $C$, $0 \leq r_i \leq 1$ and $\sum_{i}^{m} r_i = 1$. Note that an activity is usually not \emph{context-consistent} when ontology's necessary constraints discussed in Section~\ref{subsec:ontologydescription} are violated.

\begin{exmp}
	Continuing Example~\ref{exmp:running}, suppose that Bob is actually running. According to the \textsc{Incremental Activity Recognition} classifier, the current probability distribution is $45\%$ cycling, $40\%$ running, $10\%$ walking and $5\%$ standing. 
	Thanks to a dedicated Web service, it is possible to know that Bob is currently in a pedestrian area of the park where bicycles are not allowed. 
	According to the ontology, cycling is not context-consistent since it should not be performed in pedestrian areas. 
	Hence, the resulting context-refined probability distribution is $73\%$ running, $18\%$ walking and $9\%$ standing.
\end{exmp}


\subsection{Prediction Confidence Evaluation}

The \textsc{Prediction Confidence Evaluation} module is in charge of using context-refined predictions to update the activity model with new labeled samples combining self-learning and active learning. Moreover, it also applies oversampling methods in a real-time fashion to further improve the recognition of minority activity classes.

\subsubsection{Semi-supervised model update}
\label{subsubsec:semi_sup_mod}
In order to update the activity model, we apply an uncertainty sampling strategy to identify in real-time the confidence level of each refined prediction~\cite{lewis1994heterogeneous}. Given a context-refined prediction $\langle r_1, r_2, \dots, r_c \rangle$, we denote $r^\star = \max_{i} r_i$ as the probability value of the most likely activity.
If $r^\star$ exceeds a threshold $\pi$, we consider the system very confident on the current classification and we update the semi-supervised activity model with a new labeled sample (self-learning). Otherwise, if $r^\star$ is below a threshold $\rho$, where $\rho < \pi$, we consider the system uncertain about the current activity being performed by the user. In this case, an active learning process is started by asking the user to provide the ground truth about the current activity (active learning), in order to update the model accordingly. 
For the sake of usability, CAVIAR presents the user with a few predefined options picked from the most probable activities.  
When $\rho < r^\star < \pi$ we do not consider the current prediction to update the semi-supervised activity model. 

\subsubsection{Incremental data balancing}
\label{subsec:SMOTE_overlap}
During everyday life some activities are performed on average with a lower frequency than others (e.g., the amount of time that a subject spends on an elevator is usually less than the time he/she spends walking). Hence, adding new labeled samples to the incremental classifier without taking into account this aspect may lead to an unbalanced classification model and subsequently to a poor recognition rate on the ``minority'' activity classes.
For this reason, our ontology also describes (using OWL2 properties) which activity classes are known to be ``minority'' according to common-sense knowledge.
We define $M$ as the set of ``minority'' activity classes according to the ontology.

We adopt the well-known SMOTE technique~\cite{chawla2002smote} in real-time to balance the activity model by generating synthetic samples of ``minority'' activity classes. In particular, when a segment $S$ labeled as $ac \in M$ is provided as a new labeled example to update the model, we create $q$ additional synthetic labeled samples using SMOTE to further improve the classifier.

\section{Experimental evaluation}
\label{sec:evaluation}
In order to evaluate our system, we developed a data collection infrastructure to acquire a real labeled dataset consisting of both inertial sensor data and context data.
Indeed, to the best of our knowledge, there is not a publicly available dataset of labeled activities which incorporates the rich contextual information we need.
In this section, we describe 
our experimental setup, the collected dataset, and the evaluation results of CAVIAR.

\subsection{Experimental setup}

In our experimental setup, users carry their smartphone in the pants' front pocket and a smartwatch on the dominant hand's wrist.
In our data collection, we used a Nexus 5x as smartphone and an LG G Watch R as smartwatch.
Dedicated applications run on the devices to continuously collect and transmit sensor measurements to a Java server which stores data in a MongoDB database. 
Both mobile devices communicate to the server every $3$ seconds inertial sensors readings collected from their accelerometer, gyroscope and magnetometer. 

Context data is acquired by the smartphone application considering built-in sensors as well as publicly available web services.
The considered built-in sensors are: the barometer to get insights about height variations, the luminosity sensor, the microphone to obtain the environment's noise level and the GPS to obtain the user's location and speed. The considered web services are the following:
\begin{itemize}
	\item \textbf{Google's Places API~\footnote{\url{https://developers.google.com/places/web-service/intro}}}: to obtain the most likely semantic places where the user is performing the current activity.
	\item \textbf{OpenWeatherMap~\footnote{\url{https://openweathermap.org/}}}: to obtain the climatic conditions (e.g., sunny, cloudy, rainy), temperature, wind speed, etc.
	\item \textbf{Bing's Traffic API~\footnote{\url{https://docs.microsoft.com/en-us/bingmaps/rest-services/traffic/}}}: to obtain nearby traffic situation like road conditions, presence of road works, presence of car accidents, etc.
	\item \textbf{Transitland\footnote{\url{https://transit.land/}}}: to obtain information about transportation routes and stops close to the user.
\end{itemize}
The application also collects temporal context like: the moment of the day (e.g., morning, afternoon, evening), the day of the week, the season, etc.
Every $5$ seconds, context data is transmitted to the server. 
\\

\begin{figure}[h!]
	\centering
	\subfloat[Smartphone interface]{\includegraphics[width=0.3\columnwidth]{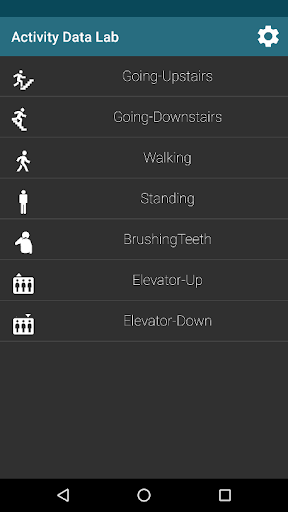}
		\label{fig:phoneapp}}
	\hfil
	\subfloat[Smartwatch interface]{\includegraphics[width=0.3\columnwidth]{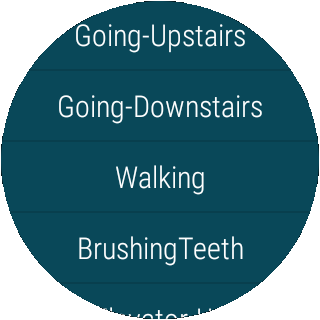}
		\label{fig:watchapp}}
	\caption{Annotation interfaces}
	\label{fig:apps}
\end{figure}

Besides data acquisition, the mobile applications have user-friendly interfaces which allow users to annotate data in real-time.
Examples of such interfaces are shown in Figure~\ref{fig:apps}. 
During the acquisition, we asked the users to use the smartwatch interface to label activities in order to spend little time in annotation and, at the same time, to make the acquisitions more realistic.

\subsection{Dataset description}

We acquired a dataset involving $26$ volunteers aged between $20$ and $28$.
The activities acquired in this dataset are the following: \emph{walking}, \emph{running}, \emph{standing}, \emph{lying}, \emph{sitting}, \emph{stairs up}, \emph{stairs down}, \emph{elevator up}, \emph{elevator down}, \emph{cycling}, \emph{moving by car}, \emph{sitting on transport}, \emph{standing on transport} and \emph{brushing teeth}.
Overall, we recorded almost $9$ hours of labeled sensor data ($\sim 350$ activity instances). Table~\ref{tab:activities} summarizes how many minutes of data we acquired for each activity. 

\begin{table}[h!]
	\centering
	\footnotesize
	\begin{tabular}{|c|c|}
		\hline
		\textbf{Activity} & \textbf{Minutes} \\
		\hline \hline
		Standing & 52 \\
		\hline
		Sitting & 56 \\
		\hline
		Lying & 40 \\
		\hline
		Walking & 96 \\
		\hline
		Running & 24 \\
		\hline
		Cycling & 24 \\
		\hline
		Brushing teeth & 16 \\
		\hline
		Stairs up & 16 \\
		\hline
		Stairs down & 16 \\
		\hline
		Elevator up & 8 \\
		\hline
		Elevator down & 16 \\
		\hline
		Sitting on transport & 60 \\
		\hline
		Standing on transport & 60 \\
		\hline
		Moving by car & 40 \\
		\hline
		\hline
		\textbf{Overall} & 524 \\
		\hline
	\end{tabular}
	
	\caption{Number of minutes acquired for each activity}
	\label{tab:activities}
	
\end{table}

Table \ref{tab:activities} shows that 
the dataset is unbalanced. As predictable, activities like \emph{taking the elevator} or \emph{brushing teeth} have been executed for a significantly shorter time than others like \emph{walking}.

The annotated sensor data has been acquired in different contexts, which include being at the office, going around in the city (Milan), driving, using public transportations, cycling and being at home. Even if the volunteers self-annotated their activities using the smartwatch, the execution of their activities was partially supervised. 
For the sake of this work, during data acquisition we were close to the volunteers to make sure that there was no technical problem with data acquisition. 

\subsection{Results}
\label{subsec:results}
In the following, we present the results of CAVIAR considering the dataset described above.
As incremental classifier, we use Online Random Forest~\cite{saffari2009line} taking advantage of the Java implementation proposed in~\cite{sztyler2017online}. 
The motivation is that Online Random Forest is the incremental version of the well-known classifier Random Forest, which proved to be one of the most effective classifiers for activity recognition~\cite{sztyler2016onbody}. 
As OWL2 reasoner we used HermiT~\cite{glimm2014hermit} in combination with the Java OWL APIs~\cite{horridge2011owl}. 

In order to evaluate the effectiveness of our technique, we also implemented two additional methods which do not rely on the \textsc{Semantic Refinement} module.
The former is called \emph{No context}, since it only considers inertial sensor data to recognize activities. 
In particular, it combines the \textsc{Incremental Activity Recognition} module (see Section~\ref{subsec:incrementalactivityrecognition}) and the \textsc{Prediction Confidence Evaluation} module (see Section~\ref{subsubsec:semi_sup_mod}) without applying our context-refinement.   

The latter method is called \emph{Context as features}. Similarly to \emph{No context}, this method does not rely on the \textsc{Semantic Refinement} module to refine activity predictions. However, this method incorporates context data directly in the feature vectors generated by the feature extraction mechanism presented in Section~\ref{subsubsec:fv}. In particular, this method extracts a) statistical features (average, variance, difference between max and min) from \emph{numeric} context data like speed or height variations and b) binary features for \emph{symbolic} context data (i.e., semantic place, weather condition, proximity to transportation routes, etc.).

We used a \emph{leave-one-subject-out} cross-validation approach to evaluate and compare CAVIAR with these two methods in terms of recognition rate and number of questions asked to the subjects.
At each fold, we apply CAVIAR to $25$ subjects to collaboratively update the activity model, which is initialized considering $1$ minute of samples for each activity. The data of the remaining subject is used to evaluate the recognition rate and the number of questions asked to the subject.

We empirically determined that the optimal window size is $w=4$, while the thresholds for semi-supervised updates are $\pi=0.7$ and $\rho=0.45$ respectively.
Table~\ref{tab:test} shows the results (in terms of overall F1 score).

\begin{table}[h!]
	\centering
	\footnotesize
	\begin{tabular}{|c|c|c|c|}
		\hline
		& \textbf{Without} & \textbf{Context} & \textbf{} \\ 
		\textbf{Activity}& \textbf{Context} & \textbf{as features} & \textbf{CAVIAR} 
		\\
		\hline \hline
		Elevator up        &0.0 				&0.04                 & \textbf{0.70}                  \\ \hline
		Elevator down      &0.02             &0.71                 & \textbf{0.83}                          \\ \hline
		Moving by car      &\textbf{0.85 }          &\textbf{0.85}               & 0.74                           \\ \hline
		Brushing teeth     &0.87           &0.91      & \textbf{0.93}                          \\ \hline
		Running            &0.97              &0.97                 & \textbf{0.99}                         \\ \hline
		Sitting            &0.96                &\textbf{0.97}     & \textbf{0.97}                           \\ \hline
		Going upstairs     &0.38          &0.69                 & \textbf{0.76}                        \\ \hline
		Going downstairs   &0.65       &0.87                 & \textbf{0.92}                           \\ \hline
		Cycling            &\textbf{0.97}  &0.90                 & 0.89                          \\ \hline
		Standing           &0.86             &0.93     & \textbf{0.95}                     \\ \hline
		Walking            &0.89             &0.94         &\textbf{0.95}                        \\ \hline
		Sitting transport  &0.60          &0.78       & \textbf{0.88}                         \\ \hline
		Standing transport &0.36        &\textbf{0.95}                & \textbf{0.95}                     \\ \hline
		\hline
		\textbf{Avg F1}   & 0.64             & 0.81             & \textbf{0.88}            \\ \hline
	\end{tabular}
	
	\caption{Recognition rate (F-$1$ score) of CAVIAR compared with alternative approaches}
	\label{tab:test}
	
\end{table}

\begin{figure}[h!]
	\centering
	\subfloat[Without context]{\includegraphics[width=0.49	\columnwidth]{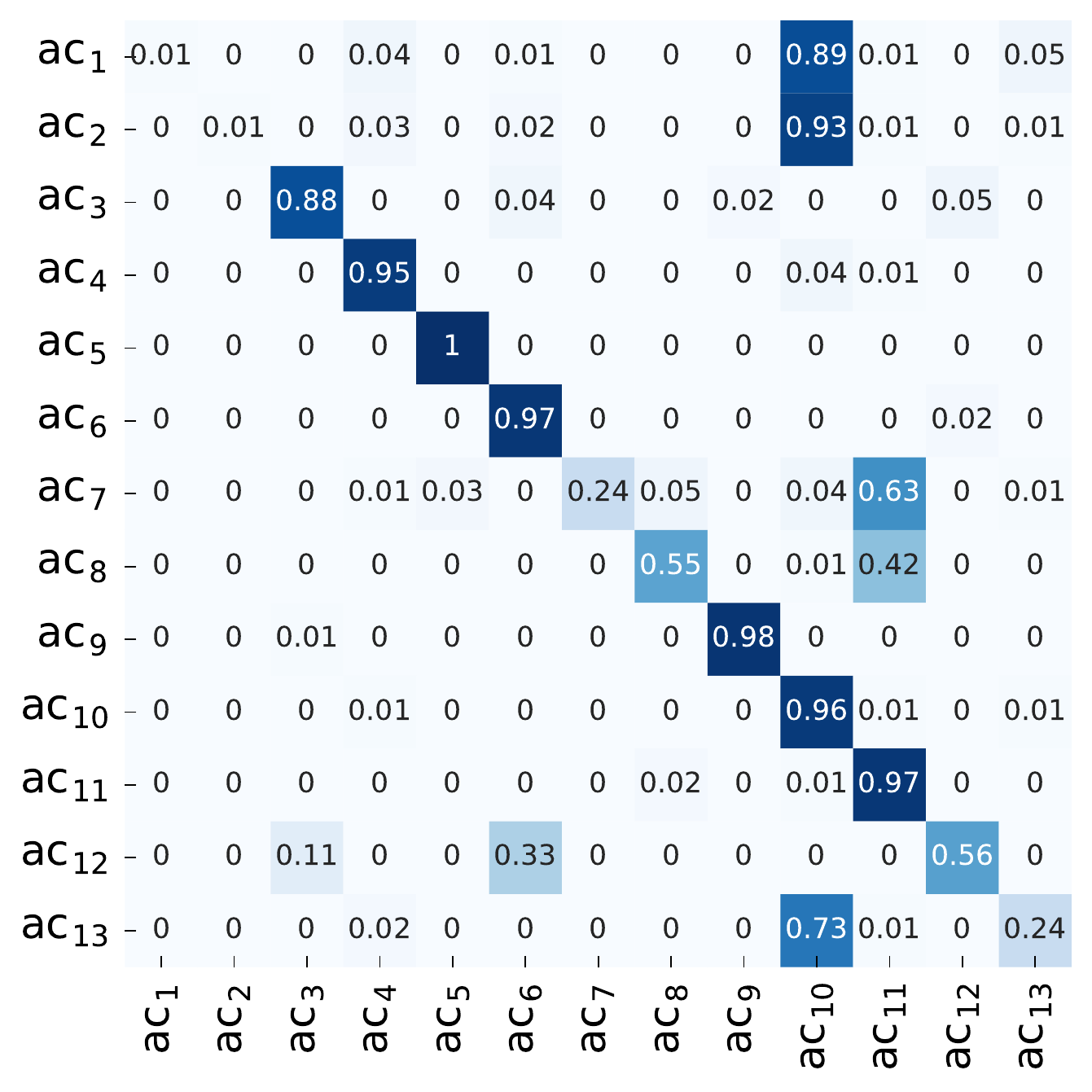}
		\label{fig:cmold}}
	\hfil
	\subfloat[CAVIAR]{\includegraphics[width=0.49\columnwidth]{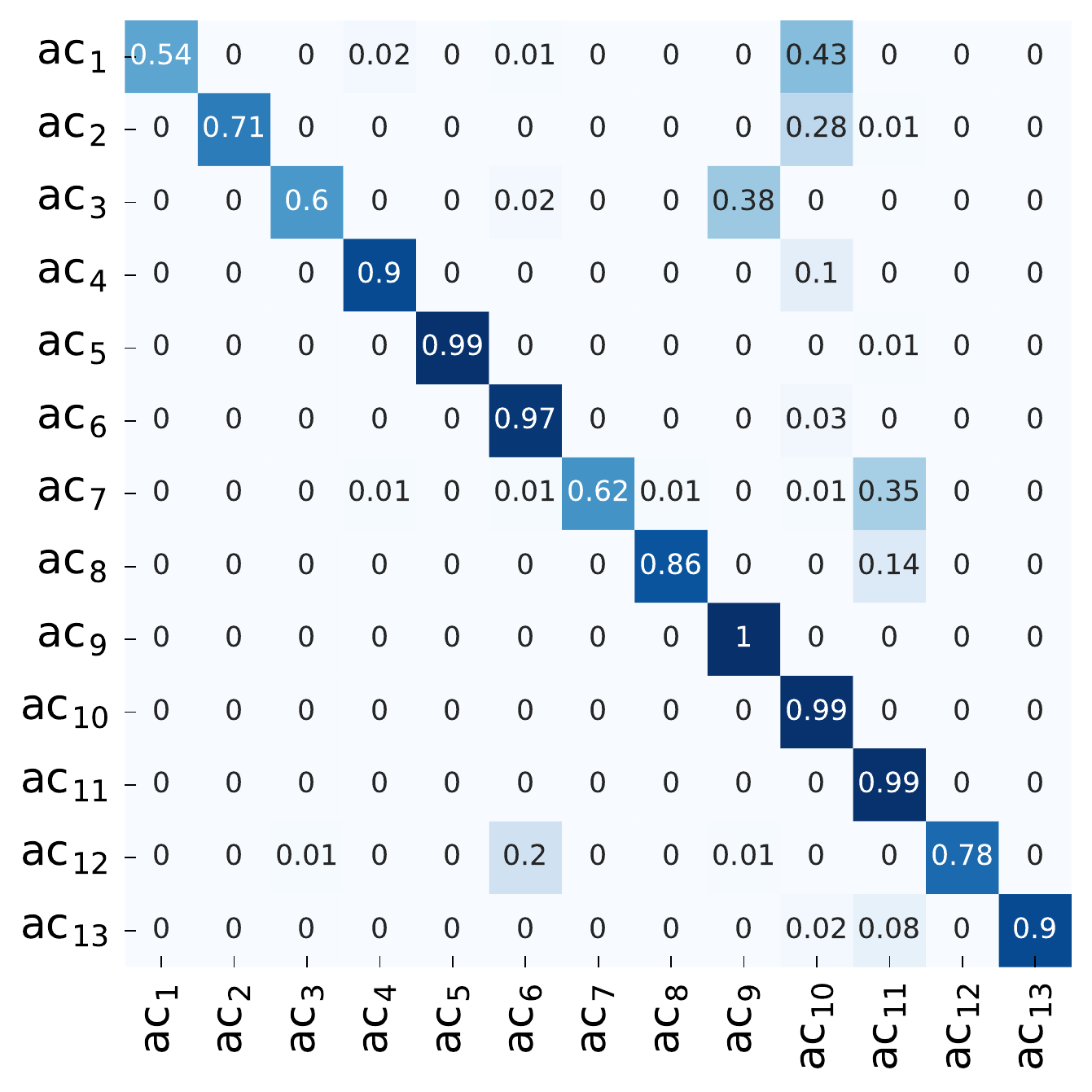}
		\label{fig:cmnew}}
	\caption{Comparison of confusion matrices. $ac_1$ = Elevator Up, $ac_2$ = Elevator Down, $ac_3$ = Moving by Car, $ac_4$ = Brushing Teeth, $ac_5$ = Running, $ac_6$ = Sitting, $ac_7$ = Going upstairs, $ac_8$ = Going Downstairs, $ac_9$ = Cycling, $ac_{10}$ = Standing, $ac_{11}$ = Walking, $ac_{12}$ = Sitting Transport, $ac_{13}$ = Standing Transport.}
	\label{fig:confusionmatrix}
\end{figure}

The results clearly show that context data 
has a significant impact on the overall recognition rate. 
Indeed, it is evident that activities like going upstairs/downstairs and sitting/standing on transport (which are more difficult to recognize only considering motion patterns) highly benefit from context data. The positive impact of context in reducing confusion between activities is also notable in the confusion matrices reported in Figure~\ref{fig:confusionmatrix}.

From the confusion matrix we  see, for example, that our approach allows the classifier to recognize the \textit{elevator up} activity, while the statistical methods confuse that activity with \emph{standing}, since they have very similar motion patterns.
In general, the fact that CAVIAR outperforms the \textit{Context as features} approach shows the value of context reasoning with common knowledge with respect to using raw context data in a statistical approach.

On the negative side, we observe that the recognition rate of CAVIAR on the \emph{moving by car} activity is lower than the ones obtained by the other approaches. 
Indeed, as Figure~\ref{fig:confusionmatrix} shows, this activity is often confused by CAVIAR with \emph{cycling}.
This is due to the fact that the context data we have to characterize those activities is similar (e.g., they are both performed outdoor in the city traffic, with a variable speed, etc.).
Hence, the semi-supervised model updates may propagate mis-predictions when the classifier has few examples of those activities. 

\begin{figure}[h!]
	\centering
	\includegraphics[width=0.8\columnwidth]{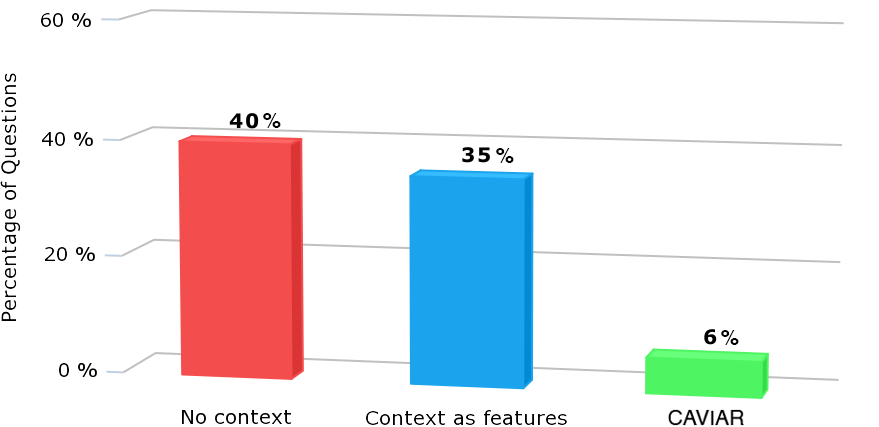}
	\caption{Percentage of triggered queries of CAVIAR compared with alternative approaches}
	\label{fig:orfquestions}
\end{figure}

Besides recognition rate, a crucial evaluation parameter is the number of questions triggered by the system, since it has a significant impact on usability. 
As Figure \ref{fig:orfquestions} shows, CAVIAR generates a significantly lower number of questions ($6\%$) with respect to \emph{No context} ($40\%$) and \emph{Context as features} ($35\%$).


Indeed, our semantic refinement technique exploits the ontology to remove from the prediction unlikely activities. This increases the confidence of the remaining activities and, at the same time, triggers our semi-supervised technique to update the activity model without bothering the user.
Results indicate that the resulting system should provide a much better user experience by limiting the number of times a user is interrupted with a question. 

\begin{figure}[h!]
	\centering
	\subfloat[F1 score]{{\includegraphics[width=5.5cm]{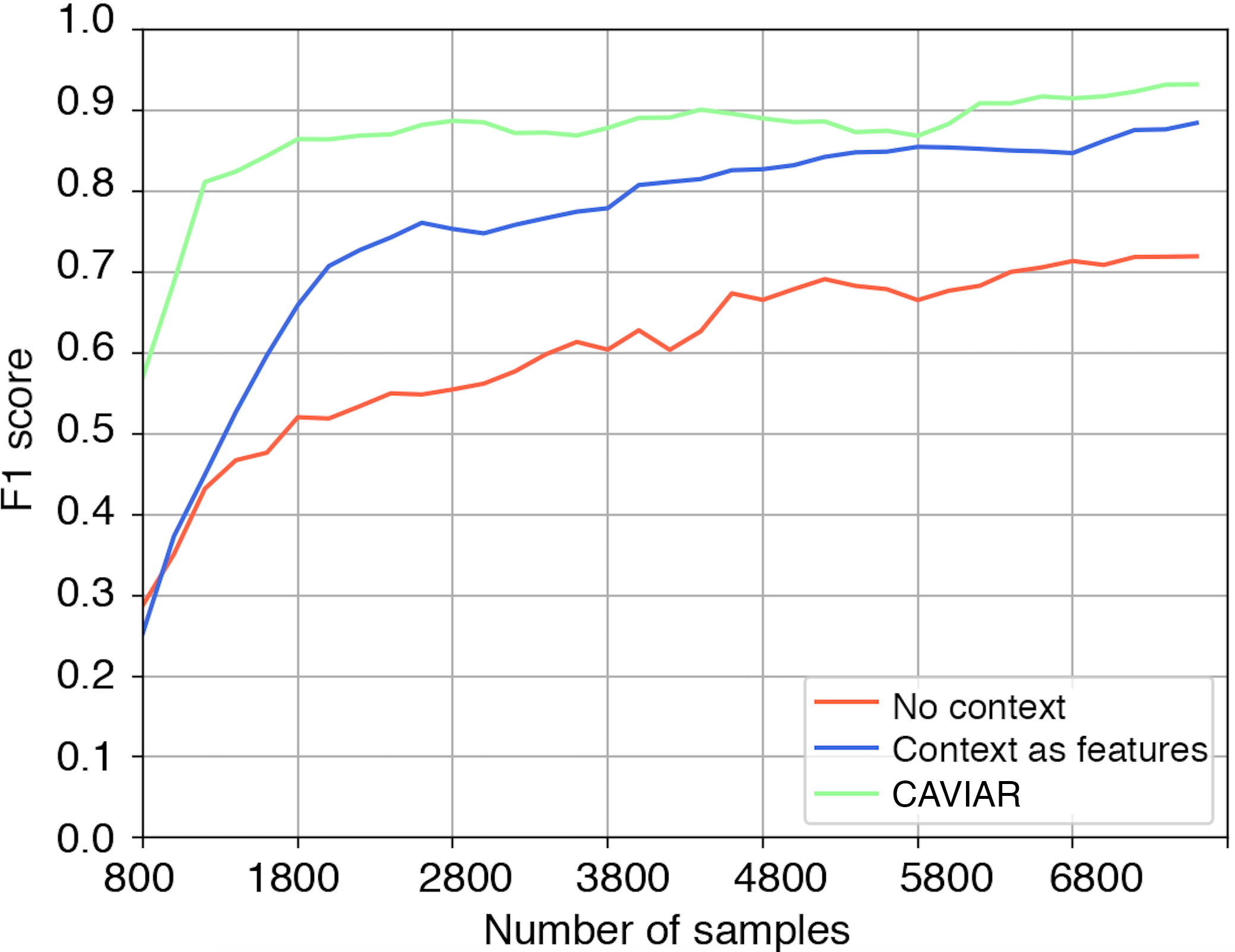} }}
	\qquad
	\subfloat[percentage of questions]{{\includegraphics[width=5.6cm]{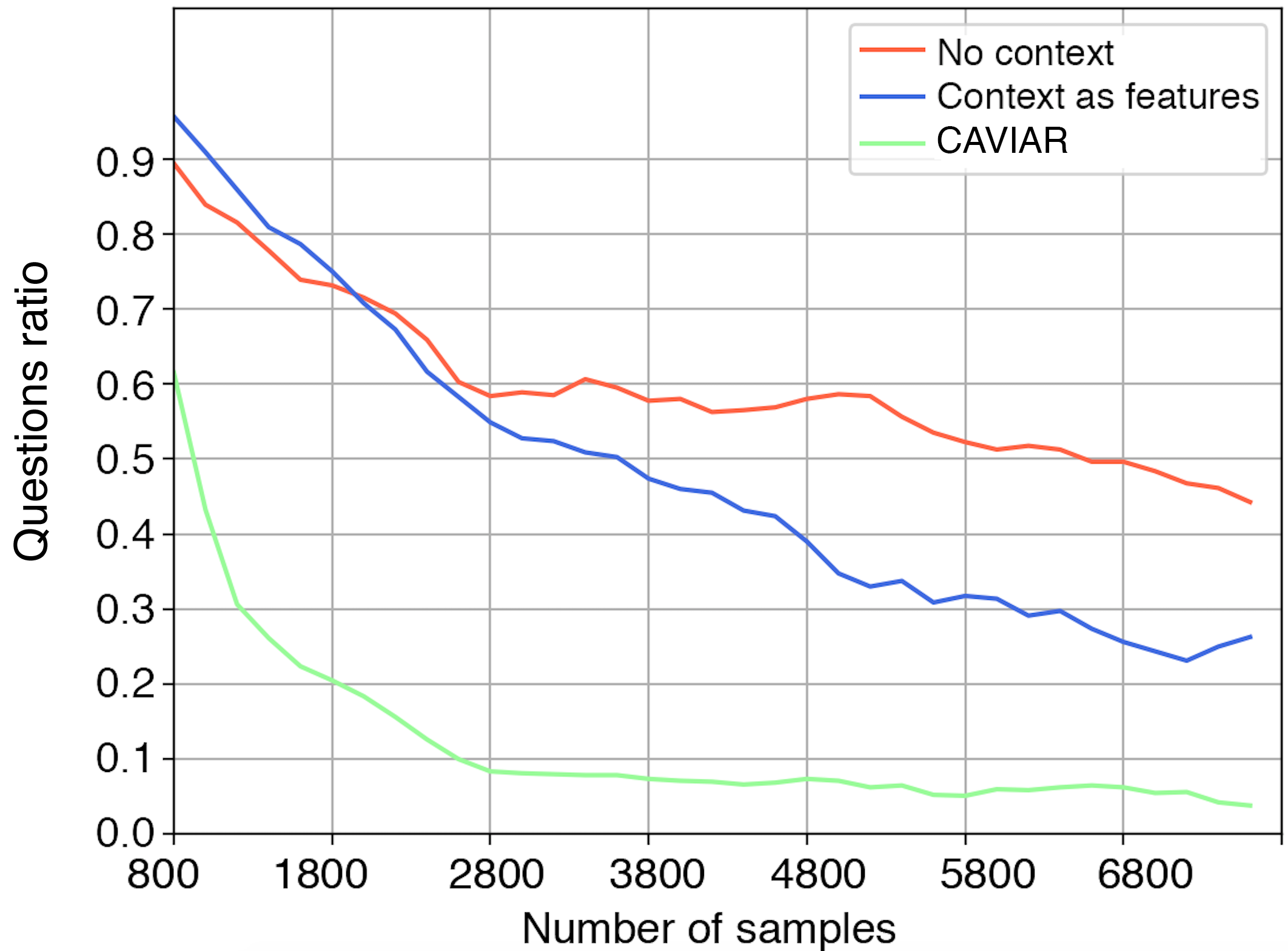} }}%
	\caption{Evolution of the recognition model over time.
		Considered activities: Running, Sitting, Cycling, Standing, Walking, Elevator up, Elevator down, Going Upstairs, Going Downstairs, Brushing Teeth, Moving by car, Sitting transport, Standing transport}%
	\label{fig:expanded}
\end{figure}

In order to evaluate how the recognition rate and the number of triggered questions evolve over time, we used the method proposed in~\cite{gama2013evaluating}.
First, we initialize the model as described above, considering $1$ minute of samples for each activity.
Then, for each sample of the dataset (considering all $26$ subjects), we use the current recognition model 
to classify it and, depending on the prediction's confidence, to update the model. 
The classification's output (i.e., the resulting most likely activity) and the corresponding ground truth are stored to evaluate the recognition rate.
In particular, we use a sliding window of $800$ samples with an overlap of $75$\% to periodically compute the overall F-$1$ score and the percentage of questions triggered to the users. Figure~\ref{fig:expanded} shows the evolution of the F-$1$ score and the number of questions of CAVIAR with respect to the other two considered methods.
It emerges that, with respect to \emph{No context} and \emph{Context as features} approaches, CAVIAR quickly reaches high recognition rates and a significant lower number of questions.

\begin{figure}[h!]
	\centering
	\subfloat[F1 score]{{\includegraphics[width=5.5cm]{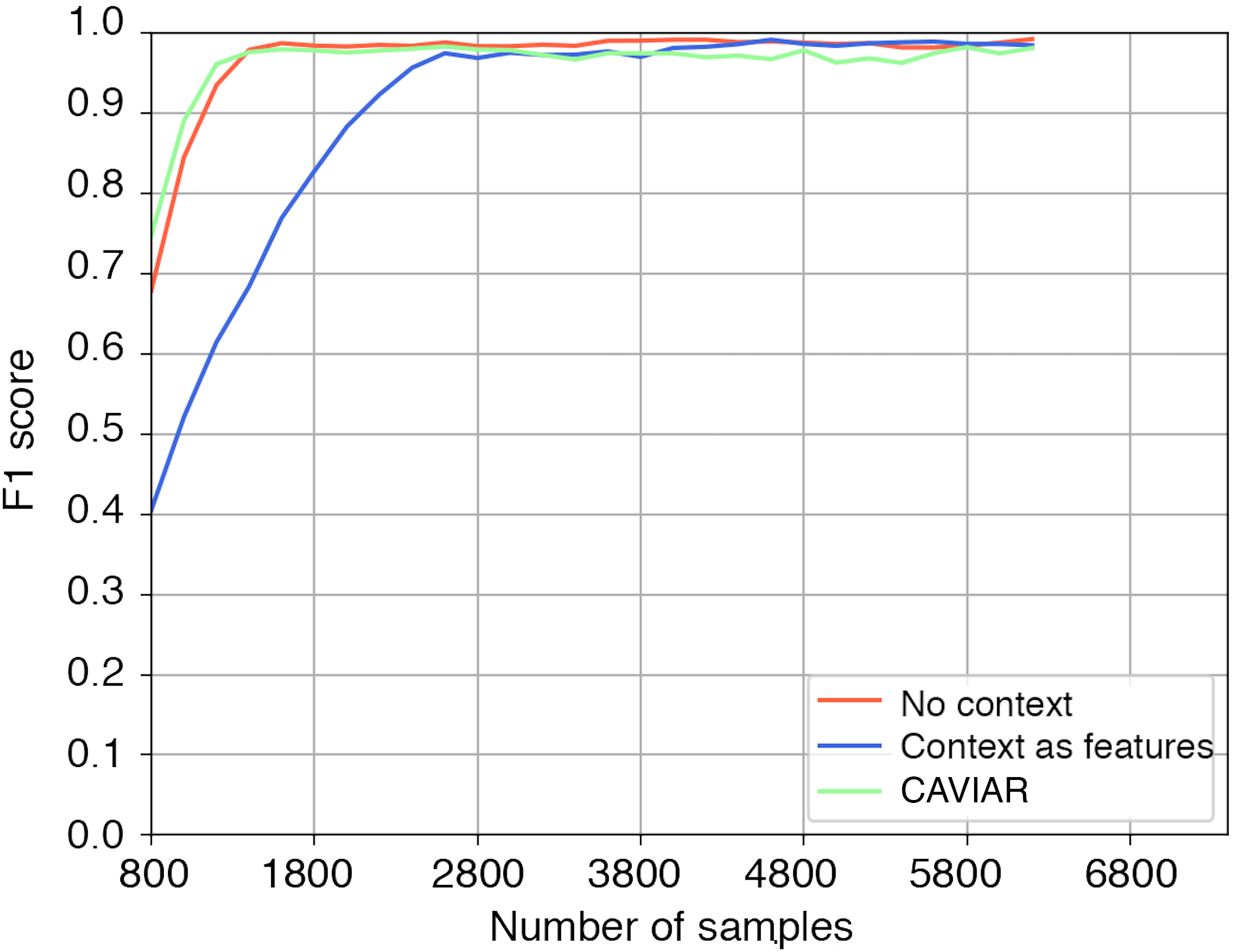} }}
	\qquad
	\subfloat[percentage of questions]{{\includegraphics[width=5.6cm]{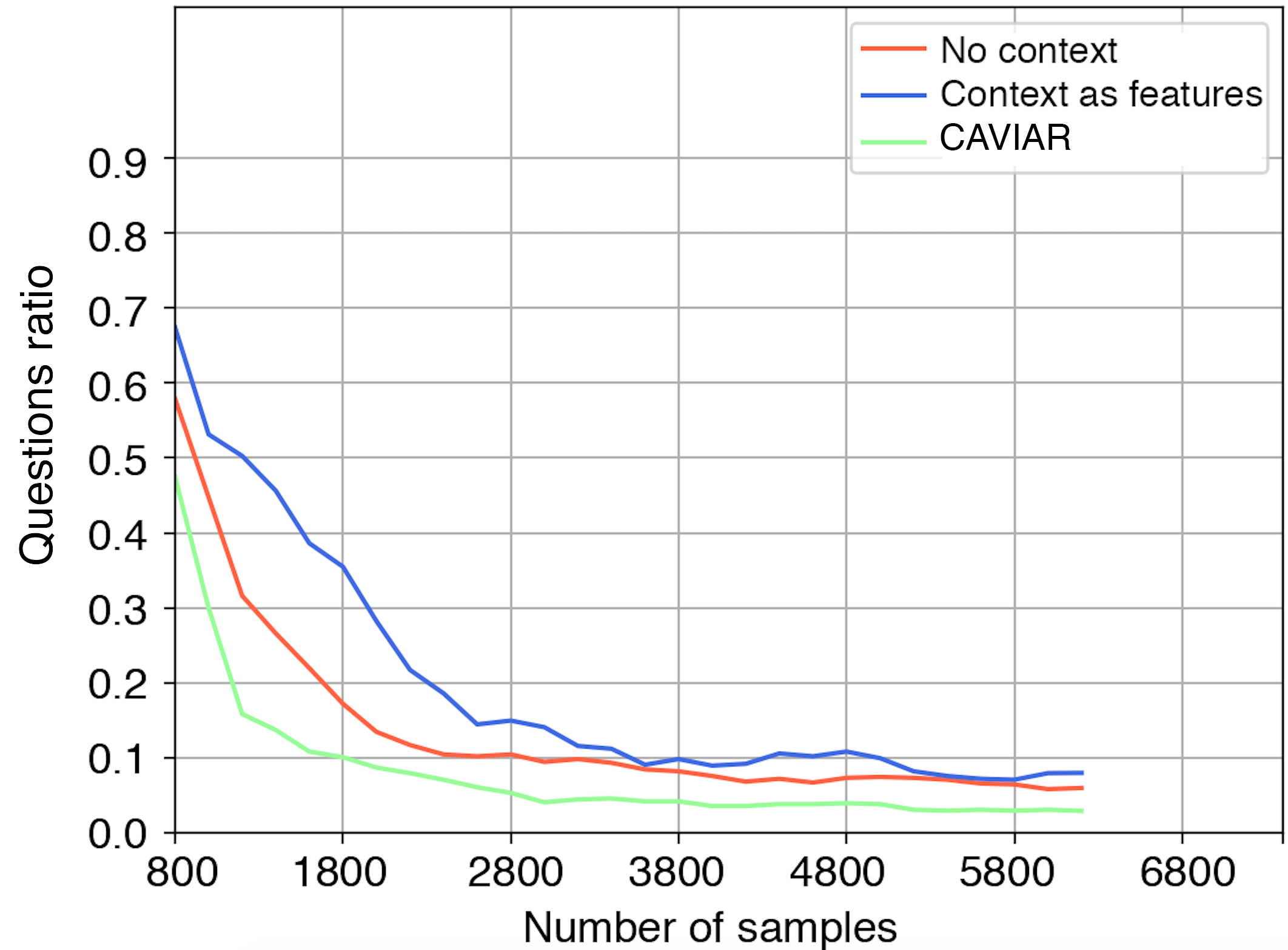} }}%
	\caption{Evolution of the recognition model over time.
		Considered activities: Running, Sitting, Cycling, Standing, Walking}%
	\label{fig:reduced}
\end{figure}

In order to further show the impact of context reasoning on activity recognition, we also evaluated our system considering different sets of activities.
Figure~\ref{fig:reduced} shows how our system performs on a restricted set of simple physical activities which are considered in the majority of related works.
We observe that those activities are poorly characterized by the context which surrounds the user, while their motion patterns can be easily discriminated by purely statistical models.

Indeed, the results show that the recognition rate and the number of questions reached by CAVIAR are similar to the ones obtained by the \emph{No Context} approach.
It also strikes out that \textit{Context as features} method has a slower learning improvement, due to the fact that additional context features add complexity to the semi-supervised activity model. 

\begin{figure}[h!]
	\centering
	\subfloat[F1 score]{{\includegraphics[width=5.5cm]{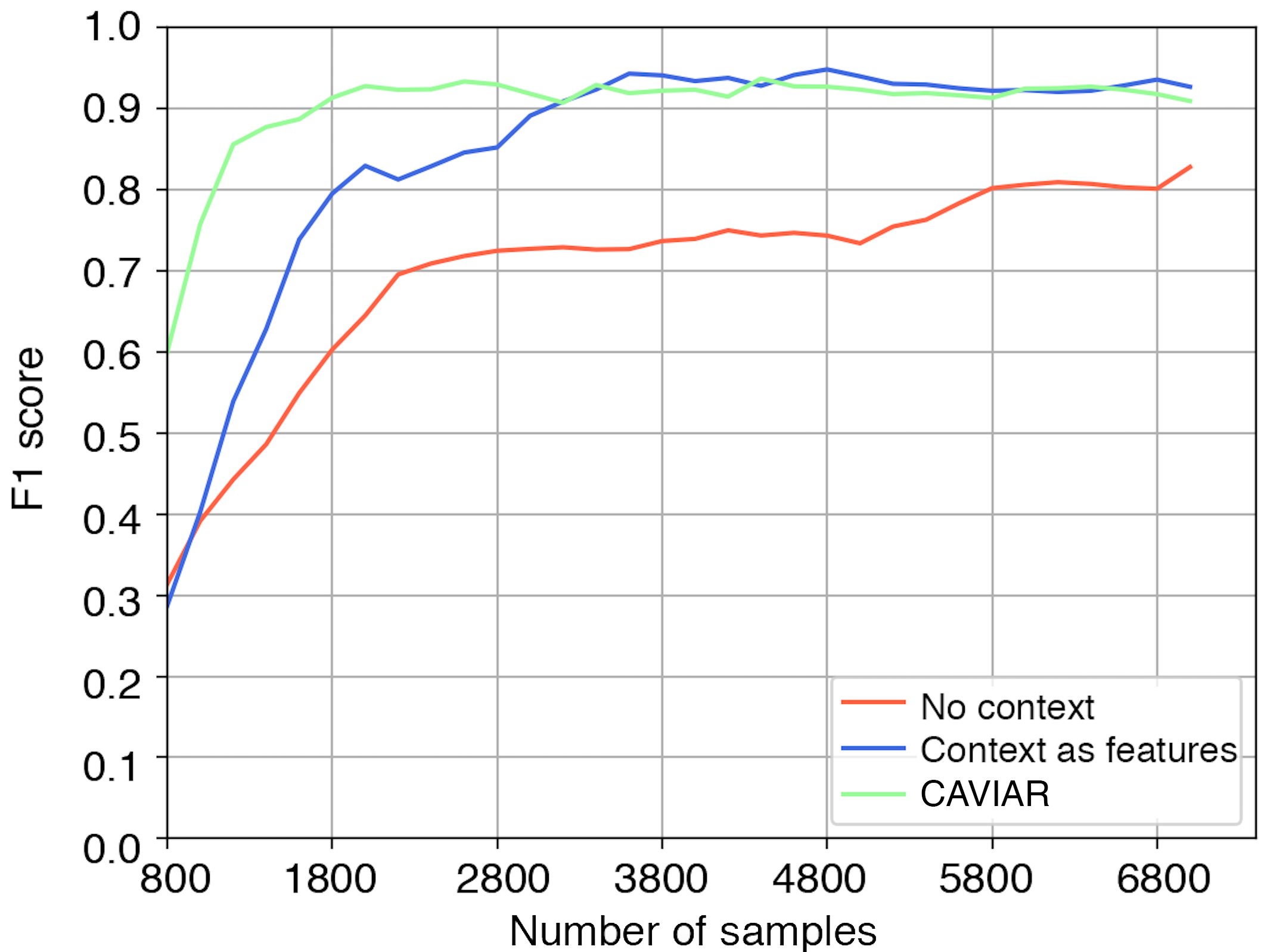} }}
	\qquad
	\subfloat[percentage of questions]{{\includegraphics[width=5.5cm]{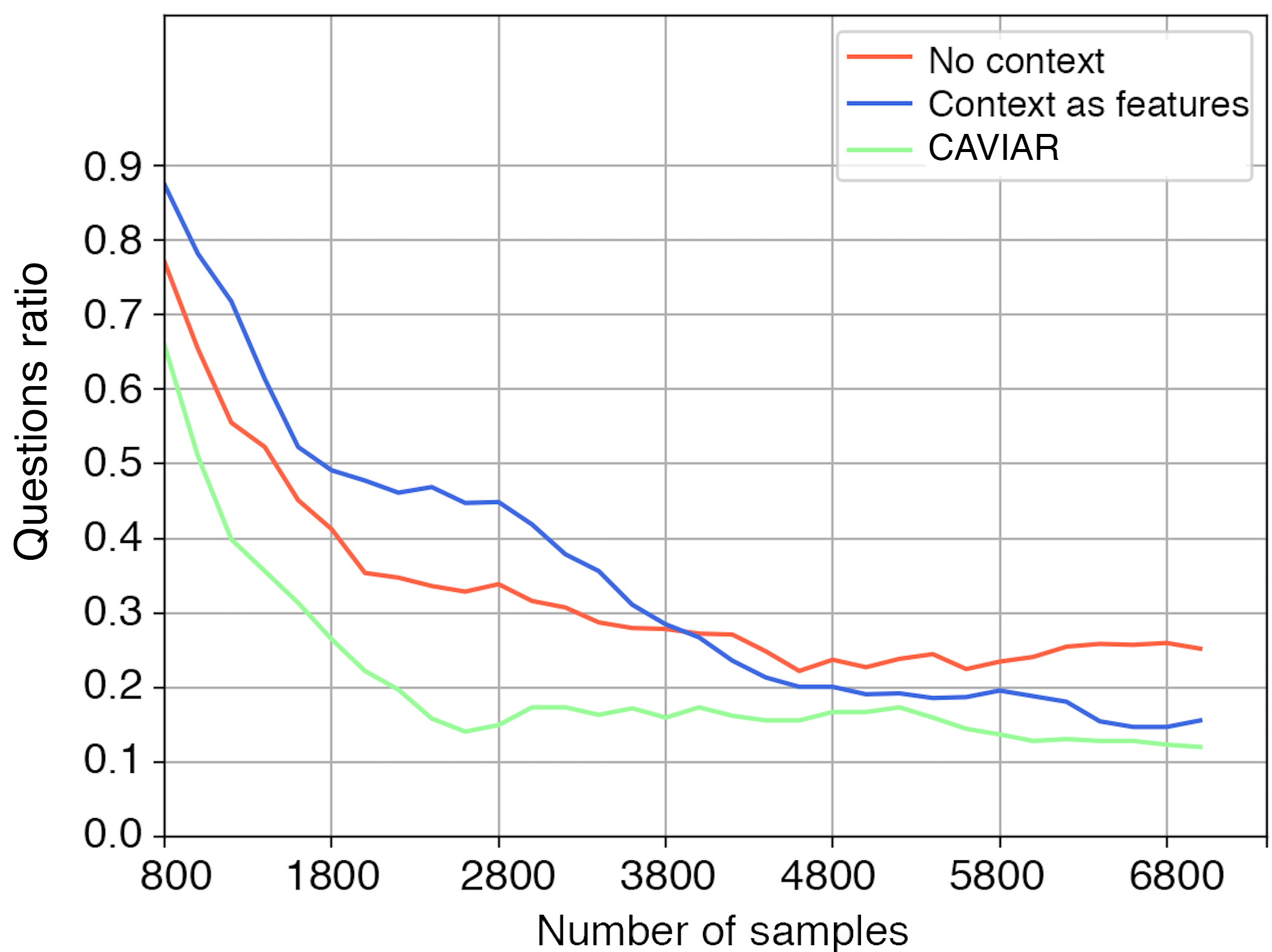} }}
	\caption{Evolution of the recognition model over time.
		Considered activities: Running, Sitting, Cycling, Standing, Walking, Elevator, Stairs, Brushing Teeth, Moving by car}
	\label{fig:classic}
\end{figure}

Finally, we also considered an ``intermediate'' set of activities. That set includes more context-dependent activities, like \textit{moving by car}, \textit{brushing teeth}, \textit{elevator} and \textit{stairs}. The results are shown in Figure \ref{fig:classic}. 

From the plots it emerges that context-data has a high impact both on the recognition rate and on the number of questions.
Indeed, the \emph{No context} method struggles to reach the performance of context-based solutions.
Moreover, CAVIAR reaches high recognition rates and a low number of questions very quickly with respect to the \textit{context as features} approach.
Hence, it emerges that considering context-data allows to significantly expand the set of considered activities.

Combining these results with the ones showed in Figure~\ref{fig:expanded}, it emerges that CAVIAR maintains the same trend independently from the considered set of activities. On the other hand, \emph{Context as Features} method's performance degrades when the complexity of the considered activities increases.

These results confirm that CAVIAR is scalable with respect to the set of considered activities and that it is effective in significantly reducing the number of triggered questions.

\section{Discussion}
\label{sec:discussion}
\subsection{Context-data as features in machine learning versus knowledge-based context reasoning}

From the results, it emerges that ontological reasoning on context-data is effective in improving the recognition rate of a semi-supervised classifier. 
However, one may point out that the performances in terms of F1 score are not too distant from the ones obtained using context-data directly as features.
It is important to note that the main advantage of our knowledge-based method is the drastically reduced number of queries which are needed to reach those high recognition rates. This aspect has a significant impact on usability in a realistic scenario. 

We also believe that our approach is significantly more scalable. One aspect to consider is that not every context source may be continuously available at the same time. Using context as features 
may lead to
missing values in the feature vectors that can potentially be used to update the classifier, which in turn may negatively impact the recognition rate. 
Another aspect is whether it becomes necessary to include additional context data while the system is running. 
In the case of using context as features, there is the need for re-training the model from scratch with new labeled data. In our method, considering new context data simply means extending the ontology. Moreover, the ontology can also be periodically revised and improved.

\subsection{Knowledge-engineering effort}

The knowledge-engineering effort required to design a comprehensive ontology is very high. Indeed, this is a task which needs to be performed by a team of domain experts and knowledge engineers. 
However, the problem is mitigated by
re-using existing ontologies.  For instance, in this work we extended the ontology presented in~\cite{RiboniB11}. 

We also point out that human modeling of knowledge about contexts is also likely to be incomplete, since it is hard to model all the possible contexts in which activities are executed. To this end, 
a promising research direction is 
to exploit semi-supervised methods to continuously refine the ontology, and at the same time learn personalized contexts. Some preliminary results in this direction have been obtained for smart-home ADLs recognition~\cite{nectar}. However, applying this concept to CAVIAR is still an open and challenging issue.

%
%

\section{Conclusion and future work}
\label{sec:conclusion}
In this paper we proposed a novel method based on the combination of context-aware reasoning and semi-supervised learning for activity recognition.
Our approach relies on ontological reasoning over context data and activities to continuously improve an incremental classifier in a semi-supervised fashion. 
We evaluated our method on a novel and rich dataset, showing the positive impact of context reasoning both on the recognition and on the number of queries triggered to the users.

A major limitation of our approach is the rigid formalism for semantic reasoning, which does not take into account the intrinsic uncertainty and incompleteness of common knowledge.
and sensor-based systems. 
Hence, in future work we plan to evaluate alternative probabilistic formalisms to model and reason about context. 
Another interesting extension may be to consider temporal sequences of activities and contexts for enhancing the semantic refinement method. This implies introducing a form of temporal reasoning.

Finally, CAVIAR could be extended to create a personalized activity model for each user.
Indeed, different subjects may have different physical characteristics and habits. This implies that there is a high variance of activity execution modalities and contexts among different users. 
The extension requires studying how to personalize the recognition model for each user and, at the same time, how to learn personalized contexts in order to further enhance the recognition rate. 


\bibliographystyle{elsarticle-num}
\section*{References}

\bibliography{references}

\begin{thebibliography}{10}
\expandafter\ifx\csname url\endcsname\relax
  \def\url#1{\texttt{#1}}\fi
\expandafter\ifx\csname urlprefix\endcsname\relax\def\urlprefix{URL }\fi
\expandafter\ifx\csname href\endcsname\relax
  \def\href#1#2{#2} \def\path#1{#1}\fi

\bibitem{lara2013survey}
O.~D. Lara, M.~A. Labrador, et~al., A survey on human activity recognition
  using wearable sensors., IEEE Communications Surveys and Tutorials 15~(3)
  (2013) 1192--1209.

\bibitem{kwapisz2011activity}
J.~R. Kwapisz, G.~M. Weiss, S.~A. Moore, Activity recognition using cell phone
  accelerometers, ACM SigKDD Explorations Newsletter 12~(2) (2011) 74--82.

\bibitem{abdallah2018activity}
Z.~S. Abdallah, M.~M. Gaber, B.~Srinivasan, S.~Krishnaswamy, Activity
  recognition with evolving data streams: A review, ACM Computing Surveys
  (CSUR) 51~(4) (2018) 71.

\bibitem{hossain2017active}
H.~S. Hossain, M.~A. A.~H. Khan, N.~Roy, Active learning enabled activity
  recognition, Pervasive and Mobile Computing 38 (2017) 312--330.

\bibitem{abdallah2015adaptive}
Z.~S. Abdallah, M.~M. Gaber, B.~Srinivasan, S.~Krishnaswamy, Adaptive mobile
  activity recognition system with evolving data streams, Neurocomputing 150
  (2015) 304--317.

\bibitem{longstaff2010improving}
B.~Longstaff, S.~Reddy, D.~Estrin, Improving activity classification for health
  applications on mobile devices using active and semi-supervised learning, in:
  Pervasive Computing Technologies for Healthcare (PervasiveHealth), 2010 4th
  International Conference on Pervasive Computing Technologies for Healthcare,
  IEEE, 2010, pp. 1--7.

\bibitem{stikic2008exploring}
M.~Stikic, K.~Van~Laerhoven, B.~Schiele, Exploring semi-supervised and active
  learning for activity recognition, in: 2008 12th IEEE International Symposium
  on Wearable Computers, IEEE, 2008, pp. 81--88.

\bibitem{liao2006location}
L.~Liao, D.~Fox, H.~Kautz, Location-based activity recognition, in: Advances in
  Neural Information Processing Systems, 2006, pp. 787--794.

\bibitem{RiboniB11}
D.~Riboni, C.~Bettini, {COSAR:} {H}ybrid reasoning for context-aware activity
  recognition, Personal and Ubiquitous Computing 15~(3) (2011) 271--289.

\bibitem{shoaib2015survey}
M.~Shoaib, S.~Bosch, O.~D. Incel, H.~Scholten, P.~J. Havinga, A survey of
  online activity recognition using mobile phones, Sensors 15~(1) (2015)
  2059--2085.

\bibitem{gyorbiro2009activity}
N.~Gy{\"o}rb{\'\i}r{\'o}, {\'A}.~F{\'a}bi{\'a}n, G.~Hom{\'a}nyi, An activity
  recognition system for mobile phones, Mobile Networks and Applications 14~(1)
  (2009) 82--91.

\bibitem{sun2010activity}
L.~Sun, D.~Zhang, B.~Li, B.~Guo, S.~Li, Activity recognition on an
  accelerometer embedded mobile phone with varying positions and orientations,
  in: International conference on ubiquitous intelligence and computing,
  Springer, 2010, pp. 548--562.

\bibitem{bao2004activity}
L.~Bao, S.~S. Intille, Activity recognition from user-annotated acceleration
  data, in: Pervasive Computing: Second International Conference, PERVASIVE
  2004, Linz/Vienna, Austria, April 21-23, 2004. Proceedings, Springer, Berlin,
  Heidelberg, 2004, pp. 1--17.

\bibitem{BullingBS14}
A.~Bulling, U.~Blanke, B.~Schiele, A tutorial on human activity recognition
  using body-worn inertial sensors, {ACM} Computing Surveys 46~(3) (2014)
  33:1--33:33.

\bibitem{CookFK13}
D.~J. Cook, K.~D. Feuz, N.~C. Krishnan, Transfer learning for activity
  recognition: A survey, Knowledge and Information Systems 36~(3) (2013)
  537--556.

\bibitem{kwon2014unsupervised}
Y.~Kwon, K.~Kang, C.~Bae, Unsupervised learning for human activity recognition
  using smartphone sensors, Expert Systems with Applications 41~(14) (2014)
  6067--6074.

\bibitem{trabelsi2013unsupervised}
D.~Trabelsi, S.~Mohammed, F.~Chamroukhi, L.~Oukhellou, Y.~Amirat, An
  unsupervised approach for automatic activity recognition based on hidden
  markov model regression, IEEE Transactions on Automation Science and
  Engineering 10~(3) (2013) 829--835.

\bibitem{lee2009unsupervised}
M.-S. Lee, J.-G. Lim, K.-R. Park, D.-S. Kwon, Unsupervised clustering for
  abnormality detection based on the tri-axial accelerometer, ICCAS-SICE 2009
  (2009) 134--137.

\bibitem{chen2014ontology}
L.~Chen, C.~D. Nugent, G.~Okeyo, An ontology-based hybrid approach to activity
  modeling for smart homes., IEEE Trans. Human-Machine Systems 44~(1) (2014)
  92--105.

\bibitem{guan2007activity}
D.~Guan, W.~Yuan, Y.-K. Lee, A.~Gavrilov, S.~Lee, Activity recognition based on
  semi-supervised learning, in: Embedded and Real-Time Computing Systems and
  Applications, 2007. RTCSA 2007. 13th IEEE International Conference on, IEEE,
  2007, pp. 469--475.

\bibitem{lee2014activity}
Y.-S. Lee, S.-B. Cho, Activity recognition with android phone using
  mixture-of-experts co-trained with labeled and unlabeled data, Neurocomputing
  126 (2014) 106--115.

\bibitem{hoque2012aalo}
E.~Hoque, J.~Stankovic, Aalo: Activity recognition in smart homes using active
  learning in the presence of overlapped activities, in: Pervasive Computing
  Technologies for Healthcare (PervasiveHealth), 2012 6th International
  Conference on, IEEE, 2012, pp. 139--146.

\bibitem{miu2015bootstrapping}
T.~Miu, P.~Missier, T.~Pl{\"o}tz, Bootstrapping personalised human activity
  recognition models using online active learning, in: 2015 IEEE International
  Conference on Computer and Information Technology; Ubiquitous Computing and
  Communications; Dependable, Autonomic and Secure Computing; Pervasive
  Intelligence and Computing, IEEE, 2015, pp. 1138--1147.

\bibitem{huynh2006towards}
T.~Huynh, B.~Schiele, Towards less supervision in activity recognition from
  wearable sensors, in: 2006 10th IEEE International Symposium on Wearable
  Computers, Citeseer, 2006, pp. 3--10.

\bibitem{rodriguez2014survey}
N.~D. Rodr{\'\i}guez, M.~P. Cu{\'e}llar, J.~Lilius, M.~D. Calvo-Flores, A
  survey on ontologies for human behavior recognition, ACM Computing Surveys
  (CSUR) 46~(4) (2014) 43.

\bibitem{akdemir2008ontology}
U.~Akdemir, P.~Turaga, R.~Chellappa, An ontology based approach for activity
  recognition from video, in: Proceedings of the 16th ACM international
  conference on Multimedia, ACM, 2008, pp. 709--712.

\bibitem{yurur2016context}
{\"O}.~Y{\"u}r{\"u}r, C.~H. Liu, Z.~Sheng, V.~C. Leung, W.~Moreno, K.~K. Leung,
  Context-awareness for mobile sensing: A survey and future directions, IEEE
  Communications Surveys \& Tutorials 18~(1) (2016) 68--93.

\bibitem{saguna2013complex}
S.~Saguna, A.~Zaslavsky, D.~Chakraborty, Complex activity recognition using
  context-driven activity theory and activity signatures, ACM Transactions on
  Computer-Human Interaction (TOCHI) 20~(6) (2013) 32.

\bibitem{PMCsurvey09}
C.~Bettini, O.~Brdiczka, K.~Henricksen, J.~Indulska, D.~Nicklas,
  A.~Ranganathan, D.~Riboni, A survey of context modelling and reasoning
  techniques, Pervasive and Mobile Computing 6~(2) (2010) 161--180.

\bibitem{banos2014window}
O.~Banos, J.-M. Galvez, M.~Damas, H.~Pomares, I.~Rojas, Window size impact in
  human activity recognition, Sensors 14~(4) (2014) 6474--6499.

\bibitem{guyon2006introduction}
I.~Guyon, A.~Elisseeff, An introduction to feature extraction, in: Feature
  extraction, Springer, 2006, pp. 1--25.

\bibitem{lewis1994heterogeneous}
D.~D. Lewis, J.~Catlett, Heterogeneous uncertainty sampling for supervised
  learning, in: Machine Learning Proceedings 1994, Elsevier, 1994, pp.
  148--156.

\bibitem{chawla2002smote}
N.~V. Chawla, K.~W. Bowyer, L.~O. Hall, W.~P. Kegelmeyer, Smote: synthetic
  minority over-sampling technique, Journal of artificial intelligence research
  16 (2002) 321--357.

\bibitem{saffari2009line}
A.~Saffari, C.~Leistner, J.~Santner, M.~Godec, H.~Bischof, On-line random
  forests, in: Computer Vision Workshops (ICCV Workshops), 2009 IEEE 12th
  International Conference on, IEEE, 2009, pp. 1393--1400.

\bibitem{sztyler2017online}
T.~Sztyler, H.~Stuckenschmidt, Online personalization of cross-subjects based
  activity recognition models on wearable devices, in: Pervasive Computing and
  Communications (PerCom), 2017 IEEE International Conference on, IEEE, 2017,
  pp. 180--189.

\bibitem{sztyler2016onbody}
T.~Sztyler, H.~Stuckenschmidt, On-body localization of wearable devices: An
  investigation of position-aware activity recognition, in: 2016 IEEE
  International Conference on Pervasive Computing and Communications (PerCom),
  IEEE Computer Society, Washington, D.C., 2016, pp. 1--9.

\bibitem{glimm2014hermit}
B.~Glimm, I.~Horrocks, B.~Motik, G.~Stoilos, Z.~Wang, Hermit: an owl 2
  reasoner, Journal of Automated Reasoning 53~(3) (2014) 245--269.

\bibitem{horridge2011owl}
M.~Horridge, S.~Bechhofer, The owl api: A java api for owl ontologies, Semantic
  Web 2~(1) (2011) 11--21.

\bibitem{gama2013evaluating}
J.~Gama, R.~Sebasti{\~a}o, P.~P. Rodrigues, On evaluating stream learning
  algorithms, Machine learning 90~(3) (2013) 317--346.

\bibitem{nectar}
G.~Civitarese, D.~Riboni, C.~Bettini, Z.~H. Janjua, R.~Helaoui, Nectar:
  Knowledge-based collaborative active learning for activity recognition, in:
  Pervasive Computing and Communications (PerCom), 2018 IEEE International
  Conference on, IEEE, 2018.

\end{thebibliography}



\end{document}